\documentclass[10pt,twocolumn,letterpaper]{article}

\usepackage{cvpr}              
\usepackage{url}
\usepackage{graphicx}
\usepackage{booktabs}
\usepackage{multirow}
\usepackage{adjustbox}
\usepackage{algorithm}
\usepackage{algorithmicx}
\usepackage{algpseudocode}
\usepackage{threeparttable}
\usepackage{pifont}
\usepackage{array}
\usepackage{enumitem}
\usepackage{bbding}
\usepackage{fontawesome}
\usepackage{diagbox}
\usepackage{svg}
\usepackage{amssymb}
\usepackage{tabularx}
\usepackage{siunitx}
\usepackage{rotating}
\usepackage{wrapfig}

\definecolor{cvprblue}{rgb}{0.21,0.49,0.74}
\usepackage[pagebackref,breaklinks,colorlinks,allcolors=cvprblue]{hyperref}

\def\paperID{5743} 
\def\confName{CVPR}
\def\confYear{2025}

\title{AnyAttack: Towards Large-scale Self-supervised Adversarial Attacks on Vision-language Models}

\author{\textbf{Jiaming Zhang\textsuperscript{\rm 1} \quad Junhong Ye\textsuperscript{\rm 2} \quad Xingjun Ma\textsuperscript{\rm 3}\thanks{Corresponding authors} \quad Yige Li\textsuperscript{\rm 4}\footnotemark[1] \quad Yunfan Yang\textsuperscript{\rm 2}} \\
\textbf{Yunhao Chen}\textsuperscript{\rm 3} \quad \textbf{Jitao Sang\textsuperscript{\rm 2,5} \quad Dit-Yan Yeung\textsuperscript{\rm 1}} \\
\textsuperscript{\rm 1}Hong Kong University of Science and Technology \
\textsuperscript{\rm 2}Beijing Jiaotong University \\ \textsuperscript{\rm 3}Fudan University \
\textsuperscript{\rm 4}Singapore Management University \
\textsuperscript{\rm 5}Peng Cheng Laboratory\\
{\centering Project Page: \url{https://jiamingzhang94.github.io/anyattack/}}
}

\begin{document}
\maketitle

\begin{abstract}


Due to their multimodal capabilities, Vision-Language Models (VLMs) have found numerous impactful applications in real-world scenarios. However, recent studies have revealed that VLMs are vulnerable to image-based adversarial attacks. Traditional targeted adversarial attacks require specific targets and labels, limiting their real-world impact.
We present \textbf{AnyAttack}, a self-supervised framework that transcends the limitations of conventional attacks through a novel foundation model approach. By pre-training on the massive LAION-400M dataset without label supervision, AnyAttack achieves unprecedented flexibility - enabling \textbf{any} image to be transformed into an attack vector targeting \textbf{any} desired output across different VLMs.
This approach fundamentally changes the threat landscape, making adversarial capabilities accessible at an unprecedented scale. Our extensive validation across five open-source VLMs (CLIP, BLIP, BLIP2, InstructBLIP, and MiniGPT-4) demonstrates AnyAttack's effectiveness across diverse multimodal tasks. Most concerning, AnyAttack seamlessly transfers to commercial systems including Google Gemini, Claude Sonnet, Microsoft Copilot and OpenAI GPT, revealing a systemic vulnerability requiring immediate attention.

\end{abstract}

\section{Introduction}

\begin{figure}[t]
\centering
\includegraphics[width=\linewidth]{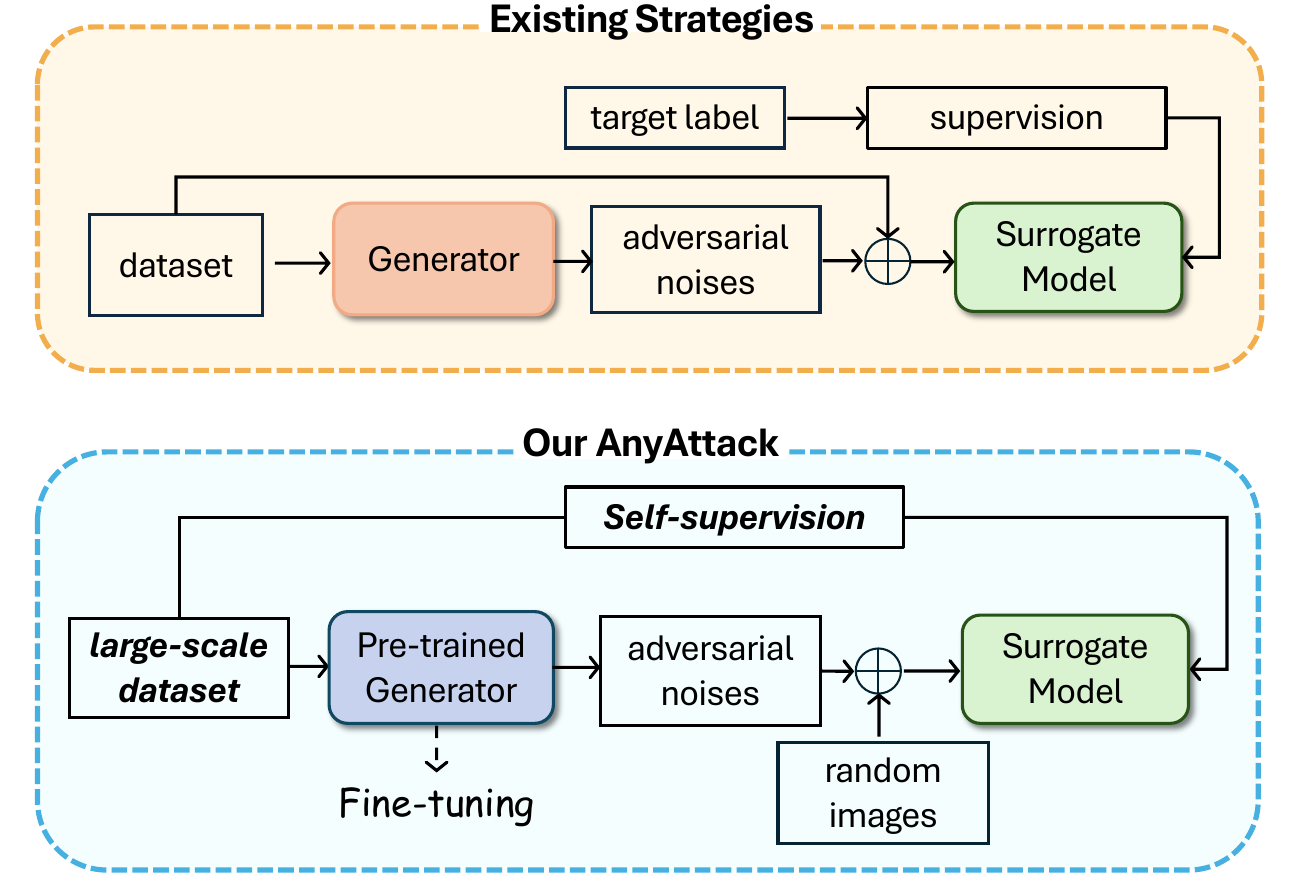} 
\caption{Comparison of existing targeted adversarial attack strategies and the our proposed self-supervised method - AnyAttack.} 
\label{fig:1} 
\vspace{-10pt}
\end{figure}

Vision-Language Models (VLMs) have exhibited remarkable performance across a diverse array of tasks, primarily attributed to the scale of training data and model size~\citep{radford2021learning, li2023blip, zhuminigpt2024}. 
Despite their remarkable performance, these models, heavily reliant on visual inputs, remain vulnerable to image-based adversarial attacks\footnote{\small For simplicity, we will refer to image-based adversarial attacks as ``adversarial attacks" in the remainder of this paper, distinguishing them from text-based adversarial attacks.}, which are carefully crafted input images designed to mislead the model into making incorrect predictions~\citep{szegedy2013intriguing}.
The evolution of adversarial attacks has progressed from general untargeted attacks (causing arbitrary errors) to more concerning targeted attacks, where adversaries can manipulate VLMs to produce specific, predetermined harmful outputs.
For instance, a benign image such as a landscape could be subtly altered to elicit harmful text descriptions such as ``violence" or ``explicit content" from the model. 
Such manipulation could have severe implications for content moderation systems, potentially leading to the removal of legitimate content or the inappropriate distribution of harmful material.

As VLMs become increasingly accessible to the public, facilitating the rapid proliferation of downstream applications, this vulnerability poses a significant threat to the reliability and security of VLMs in real-world scenarios. 
Therefore, exploring new targeted attack methods tailored to VLMs is crucial to address these vulnerabilities.
However, existing targeted attack methods on VLMs present challenges due to the reliance on target labels for supervision, which limits the scalability of the training process. 
For example, it is impractical to expect a generator trained on ImageNet~\citep{russakovsky2015imagenet} to produce effective adversarial noise for VLMs.
To overcome this limitation, we propose a novel self-supervised framework, \textbf{AnyAttack}, which leverages the original image itself as supervision, enabling \textbf{any} image to be transformed into an attack vector targeting \textbf{any} desired output across different VLMs.
Our approach involves pre-training a generator on the large-scale LAION-400M dataset~\citep{schuhmann2021laion}, enabling the pre-trained noise generator to learn comprehensive noise patterns from diverse image data.
Through self-supervised adversarial noise pre-training, we can further fine-tune the pre-trained generator on downstream datasets for adapting downstream vision-language tasks.
The large-scale pre-training establishes a foundation model, thereby enhancing the potential for developing more powerful adversarial attacks.
Our framework, to the best of our knowledge, is the first to implement the ``pre-training and fine-tuning" paradigm for targeted adversarial attacks at scale, breaking the barriers of traditional attack methods.
\cref{fig:1} highlights the distinctions between our method and existing strategies.

To demonstrate the effectiveness of our approach, we conduct extensive experiments on 5 target VLMs (CLIP~\citep{radford2021learning}, BLIP~\citep{li2022blip}, BLIP2~\citep{li2023blip}, InstructBLIP~\citep{dai2023instructblip}, and MiniGPT-4~\citep{zhuminigpt2024}), across 3 multimodal tasks (image-text retrieval, multimodal classification, and image captioning). 
We also evaluate our method on commercial VLMs, including Google Gemini, Claude Sonnet, Microsoft Copilot and OpenAI GPT.

In summary, our main contributions are:
\begin{itemize}
    \item We propose \textbf{AnyAttack}, a self-supervised framework that utilizes the original image as supervision, allowing any image to be transformed into an attack vector targeting any desired output across different VLMs.

    \item Our framework is the first to adopt the ``pre-training and fine-tuning" paradigm for targeted adversarial attacks, pre-training a noise generator on the large-scale LAION-400M dataset and fine-tuning it for downstream vision-language tasks.

    \item We demonstrate the effectiveness of our AnyAttack on five mainstream open-source VLMs across three multimodal tasks. Additionally, we successfully transfer our attack to four commercial VLMs.
\end{itemize}

\begin{figure*}[t]
\centering
\includegraphics[width=\textwidth]{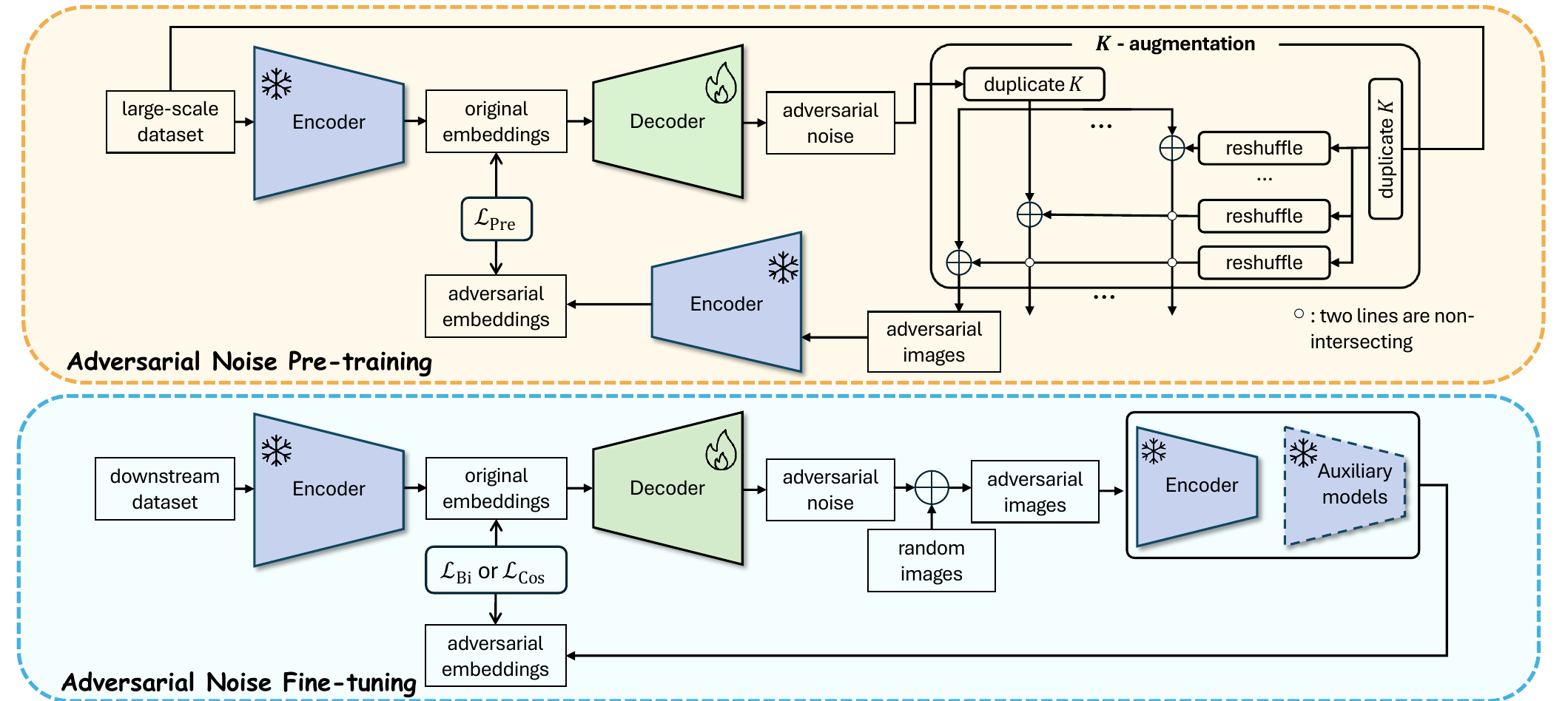} 
\caption{Overview of the proposed AnyAttack: a self-supervised framework consisting of pre-training and fine-tuning stages.} 
\label{fig:pretraining} 
\vspace{-10pt}
\end{figure*}

\section{Related Work}\label{sec:related_work}

\paragraph{Targeted Adversarial Attacks.}
A number of works have been proposed to improve the effectiveness and transferability of targeted adversarial attacks against vision models. Input augmentation techniques such as image translation \citep{dong2019evading}, cropping \citep{wei2023enhancing}, mixup \citep{wang2021admix, liu2024boosting}, and resizing \citep{xie2019improving}, have been employed to increase the diversity of adversarial input, thus improving their transferability across different target models. 
In addition, adversarial fine-tuning and model enhancement techniques have been explored to increase the attack capabilities of surrogate models \citep{springer2021little, zhang2023low, wu2024improving}. These methods typically involve retraining the surrogate models with a mix of clean and adversarial examples to improve their robustness against future attacks.
Furthermore, optimization techniques have evolved to stabilize the update processes during adversarial training. Methods such as adaptive learning rates and gradient clipping have been integrated to ensure more consistent updates and enhance the overall performance of the adversarial attacks \citep{dong2018boosting, wang2021enhancing, lin2023linnesterov}. These advancements collectively contribute to the development of more effective and transferable adversarial attacks in the realm of vision models.

\paragraph{Jailbreak Attacks on VLMs.}
Multimodal jailbreaks primarily exploit cross-modal interaction vulnerabilities in VLMs.
These attacks manipulate inputs of text~\citep{wu2023jailbreaking_text}, images~\citep{carlini2024aligned,gong2023figstep,qi2024visual,niu2024jailbreaking}, or both simultaneously~\citep{bi_modal,wang2024white}, aiming to elicit harmful but \emph{non-predefined} responses. 
In contrast, image-based adversarial attacks focus on manipulating the image encoder of VLMs. 
The objective is to induce adversary-specified, \emph{predetermined} responses through precise visual manipulations. 

\paragraph{Adversarial Attacks on VLMs.}

Adversarial research on VLMs is relatively limited compared to the extensive studies on vision models, with the majority of existing attacks focusing primarily on untargeted attacks.
Co-Attack~\citep{zhang2022towards} was among the first to perform white-box untargeted attacks on several VLMs. 
Following this, more approaches have been proposed to enhance adversarial transferability for black-box untargeted attacks~\citep{lu2023set, zhou2023advclip, yin2024vlattack, xu2024highly}.
Cross-Prompt Attack~\citep{luo2024an} investigates a novel setup for adversarial transferability based on the prompts of LLMs.
AttackVLM~\citep{zhao2024evaluating} is the most closely related work, using a combination of text inputs and popular text-to-image models to generate guided images for creating targeted adversarial images. 
Although their approach shares a similar objective with our work, our method distinguishes itself by being self-supervised and independent of any text-based guidance.

\section{Proposed Attack}

In this section, we first present the preliminaries on targeted adversarial attacks and then introduce our proposed method.

\subsection{Preliminaries and Adversary's Settings}

\paragraph{Threat Model.}
This work focuses on transfer-based black-box attacks, where the adversary generates an adversarial image \( x' \) using a fully accessible pre-trained surrogate model \( f_s \). The adversary has no knowledge of the target VLMs \( f_t \), including its architecture and parameters, nor can they leverage the outputs of \( f_t \) to reconstruct adversarial images. The adversary's objective is to cause the target VLM \( f_t \) to incorrectly match the adversarial image \( x' \) with the target text description \( y_t \).

We begin by formulating the problem of targeted adversarial attacks. Let \( f_s \) represent a pre-trained surrogate model, and  \( \mathcal{D} = \{(x, y)\} \) denote the image dataset, where \( x \) is the original image and \( y \) is the corresponding label (description). The attacker's objective is to craft an adversarial example \( x' = x + \delta \) that misleads the target model \( f_t \) into predicting a predefined target label \( y_t \). 
In the context of VLMs, this objective requires that \( x' \) aligns with \( y_t \) as a valid image-text pair. The process of generating targeted adversarial images typically involves finding a perturbation \( \delta \) using the surrogate model \( f_s \). \cref{tab:attack_formulations} highlights the key differences between our approach and current strategies, where \( x_r \) denotes an irrelevant random image. 

\begin{table}[t]
\centering
\begin{adjustbox}{width=\linewidth}
\begin{tabular}{c|l}
\toprule
\textbf{Strategy} & \textbf{Formulation} \\ 
\hline
Target label supervision & $\min \mathcal{L}(f_s(x + \delta), y_t), \quad \text{s.t.} \ y_t \neq y$ \\ 
\hline
Target image supervision & $\min \mathcal{L}(f_s(x + \delta), f_s(x_t)), \quad \text{s.t.} \ x_t \neq x$ \\ 
\hline
\textbf{AnyAttack (ours)} & $\min \mathcal{L}(f_s(\delta + x_r), f_s(x))$ \\ 
\bottomrule
\end{tabular}
\end{adjustbox}
\caption{The formulation of different attack strategies. The existing strategies rely on explicit target supervision, whereas our AnyAttack is unsupervised.}
\label{tab:attack_formulations}
\vspace{-10pt}
\end{table}

\subsection{AnyAttack}

\paragraph{Framework Overview.}
Our proposed framework, \textbf{AnyAttack}, employs two phases: \emph{self-supervised adversarial noise pre-training} and \emph{self-supervised adversarial noise fine-tuning}.
\cref{fig:pretraining} provides the self-supervised framework overview including pre-training and fine-tuning. 

For \emph{self-supervised adversarial noise pre-training}, we train a decoder \( F \), to produce adversarial noise \( \delta \) on large-scale datasets \( \mathcal{D}_p \), using frozen encoder \( E \) as the surrogate model. Given a batch of images \( x \), we extract their embeddings using a frozen image encoder \( E \). These normalized embeddings \( \mathbf{z} \) are then fed into the decoder \( F \), which generates adversarial noise \( \delta \) corresponding to the images \( x \). To enhance generalization and computational efficiency, we introduce a \( K \)-augmentation strategy that creates multiple shuffled versions of the original images within each mini-batch. During this process, adversarial noise is added to the shuffled original images (random images) to produce the adversarial images.

For \emph{self-supervised adversarial noise fine-tuning}, we adapt the pre-trained decoder \( F \) to a specific downstream dataset \( \mathcal{D}_f \). 
We use an unrelated random image \( x_r \) from an external dataset \( \mathcal{D}_e \) as the clean image to synthesize the adversarial image \( x_r + \delta \).

\paragraph{Self-supervised Adversarial Noise Pre-training} aims to train the generator on large-scale datasets, enabling it to handle a diverse array of input images as potential targets. 
Unlike existing methods, it does not require target labels or target images as supervision throughout the training process.
Our objective can be formulated as follows:
\begin{equation}
\label{eq:goal}
\min \mathcal{L}(f_s(\delta + x_r), f_s(x)), \quad \text{s.t.} \ x_r \neq x,
\end{equation}
where \( x_r \) is a random image that is unrelated to \( x \), while the adversarial noise \( \delta \) is designed to align with the original image \( x \) within the surrogate model’s embedding space, and \( \mathcal{L} \) denotes the similarity function.

Given a batch of \( n \) images \( x \in \mathbb{R}^{n \times H \times W \times 3} \) from the large-scale training dataset $\mathcal{D}_p$, we employ the CLIP ViT-B/32 image encoder, which is frozen during training, as the encoder \( E \), to obtain the normalized embeddings \( E(x) = \mathbf{z}  \in \mathbb{R}^{n \times d} \) corresponding to the original images \( x \), where \( d \) represents the embedding dimension (i.e., 512 for CLIP ViT-B/32). 
Subsequently, we deploy an initialized decoder \( F \), which maps the embeddings \( \mathbf{z} \) to adversarial noise \( D(\mathbf{z}) = \delta \in \mathbb{R}^{n \times H \times W \times 3} \) corresponding to the original images \( x \).
We expect the generated noises \( \delta \) to serve as adversarial noise representative of the original images \( x \). 
Our goal is for the generated noises \( \delta \), when added to random images \( x_r \), to be interpreted by the encoder \( E \) as the original images \( x \), i.e., \( E(x_r + \delta) = E(x) \).

To increase the number of random images within every batch, we present the \( K \)-augmentation strategy, which duplicates both adversarial noises \( \delta \) and the original images \( x \) \( K \) times, forming \( K \) mini-batches. 
For each mini-batch, the order of the adversarial noises remains consistent, while the order of the original images is shuffled within the mini-batch, referred to as shuffled images. 
These shuffled images are then added to the corresponding adversarial noise, resulting in adversarial images \( x' \).
Next, the adversarial images are fed into \( F \) to produce adversarial embeddings $\mathbf{z}^{(adv)}$, which are then used for subsequent calculations against the original embeddings $\mathbf{z}$.

Finally, we introduce the \emph{adversarial noise pre-training loss} $\mathcal{L}_{\text{Pre}}$. 
It maximizes the cosine similarity between positive sample pairs, defined by the \( i \)-th elements of adversarial and original embeddings in each mini-batch, while minimizing the similarity between the negative pairs, which consist of all other elements.
This setup creates $n$ positive pairs and $n(n-1)$ negative pairs in every mini-batch, with gradients accumulated to update $F$:
\begin{equation}
\label{eq:4}
\mathcal{L}_{\text{Pre}} = -\frac{1}{n} \sum_{i=1}^n \log \frac{ \exp \left( \mathbf{z}_i \cdot \mathbf{z}_i^{(adv)} / \tau(t) \right) }{ \sum_{j=1}^n \exp \left( \mathbf{z}_i \cdot \mathbf{z}_j^{(adv)} / \tau(t) \right) },
\end{equation}
where \( \mathbf{z}_i \) and \( \mathbf{z}_i^{(adv)} \) are the \( \ell_2 \)-normalized embeddings of the \( i \)-th sample from original images \( x \) and adversarial images \( x' \).
\( \tau(t) \) is the temperature at step \( t \), enabling the model to dynamically adjust the hardness of negative samples during training.
We set a relatively large initial temperature \( \tau_0 \) at the beginning of training and gradually decrease it, reaching the final temperature \( \tau_{\text{final}} \) after a certain number of steps \( T \):
\begin{equation}
\label{eq:5}
\tau(t) = \tau_0 \left( \frac{\tau_{\text{final}}}{\tau_0} \right)^{\frac{t}{T}} = \tau_0 \exp\left( -\lambda t \right).
\end{equation}

\paragraph{Self-supervised Adversarial Noise Fine-tuning} refines the pre-trained decoder \( F \) on downstream vision-language datasets using task-specific objective functions, facilitating its adaptation to particular domains and multimodal tasks.

Given a batch of \( n \) images \( x \in \mathbb{R}^{n \times H \times W \times 3} \) from the downstream dataset \( \mathcal{D}_f \), the encoder \( E \) remains frozen and outputs the embeddings \( \mathbf{z} \), which are then fed into the decoder \( F \) to generate the noise \( \delta \). 
Since the size of \( \mathcal{D}_f \) is much smaller than that of \( \mathcal{D}_p \), we randomly select images from an external dataset \( \mathcal{D}_e \) as random images \( x_r \), which are then added to the generated noise \( \delta \) to create adversarial images.
To improve transferability, we incorporate auxiliary models alongside the encoder \( E \), forming an ensemble surrogate.

Depending on the downstream tasks, \emph{self-supervised adversarial noise fine-tuning} employs two different fine-tuning objectives. 
The first strategy is tailored for the image-text retrieval task, which imposes stricter requirements for distinguishing between similar samples. It demands robust retrieval performance in bi-directions: from \( \mathbf{z}^{(adv)} \) to \( \mathbf{z} \) and from \( \mathbf{z} \) to \( \mathbf{z}^{(adv)} \), denoted as $\mathcal{L}_{\text{Bi}}$.
\begin{equation}
\label{eq:6}
\begin{aligned}
\mathcal{L}_{\text{Bi}} = \frac{1}{2n} \sum_{i=1}^n \Bigg( 
    & -\log \frac{ \exp ( \mathbf{z}_i \cdot \mathbf{z}_i^{(adv)} / \tau ) }
    { \sum_{j=1}^n \exp ( \mathbf{z}_i \cdot \mathbf{z}_j^{(adv)} / \tau ) } \\
    & -\log \frac{ \exp ( \mathbf{z}_i^{(adv)} \cdot \mathbf{z}_i / \tau ) }
    { \sum_{j=1}^n \exp ( \mathbf{z}_i^{(adv)} \cdot \mathbf{z}_j / \tau ) }
\Bigg).
\end{aligned}
\end{equation}

The second strategy is suited for general tasks, such as image captioning, multimodal classification, and other broad vision-language applications. It requires \( \mathbf{z}_i^{(adv)} \) to match \( \mathbf{z}_i \), so we employ cosine similarity to align \( \mathbf{z}_i^{(adv)} \) with \( \mathbf{z}_i \), denoting this objective as \( \mathcal{L}_{\text{Cos}} \).

\section{Experiments}

\begin{table*}[t]
\centering
\small
    \setlength{\aboverulesep}{0pt} 
    \setlength{\belowrulesep}{0pt}
    \setlength{\extrarowheight}{0pt} 
\begin{adjustbox}{max width=\textwidth}
\begin{tabular}{@{}clrrrrrrr@{}}
\toprule[1pt]
\multirow{2}{*}{ Model} & \multirow{2}{*}{Attack Method} & \multicolumn{7}{c}{MSCOCO} \\
\cmidrule(lr){3-9}
& & TR@1 & TR@5 & TR@10 & IR@1 & IR@5 & IR@10 & R@Mean \\
\midrule[1pt]
\multirow{10}{*}{\rotatebox{0}{\textbf{ViT-B/16}}} 
&  AttackVLM-ii & 0.4 & 1.0 & 1.4 & 0.24 & 1.08 & 2.16 & 1.05 \\
&  AttackVLM-it & 0.2 & 1.4 & 1.8 & 0.16 & 1.16 & 2.12 & 1.14 \\
& SASD-WS-Cos & 6.0 & 17.0 & 24.8 & 9.08 & 24.39 & 34.55 & 19.30 \\
& SASD-WS-MSE & 4.8 & 18.4 & 25.6 & 8.20 & 25.87 & 35.15 & 19.67 \\
& SU-Cos & 6.8 & 20.4 & 27.8 & 11.11 & 25.70 & 33.34 & 20.86 \\
& SU-MSE & 6.8 & 20.6 & 27.0 & 10.83 & 25.10 & 32.62 & 20.49 \\
\cmidrule[0.5pt]{2-9}
& \emph{AnyAttack-Cos}  & 8.6 & 21.2 & 29.6 & 10.80 & 27.59 & 37.50 & 22.55 \\
& \emph{AnyAttack-Bi}  & \underline{12.2} & \underline{26.2} & \underline{33.8} & \underline{12.63} & 31.71 & 40.86 & \underline{26.23} \\
& \emph{AnyAttack-Cos w/ Aux} &  8.4 & 24.8 & 33.0 & 11.59 & \underline{32.10} & \underline{44.98} & 25.81 \\
& \emph{AnyAttack-Bi w/ Aux} &  \textbf{14.8} & \textbf{36.8} & \textbf{48.0} & \textbf{17.59} & \textbf{42.02} & \textbf{56.05} & \textbf{35.88} \\
\midrule[1pt]

\multirow{10}{*}{\rotatebox{0}{\textbf{ViT-L/14}}} 
&  AttackVLM-ii & 0.2 & 1.0 & 1.6 & 0.24 & 0.60 & 1.32 & 0.83 \\
&  AttackVLM-it & 0.4 & 0.8 & 1.4 & 0.12 & 0.76 & 1.48 & 0.83 \\
& SASD-WS-Cos & 3.8 & 11.6 & 18.8 & 7.20 & 18.43 & 26.35 & 14.36 \\
& SASD-WS-MSE & 5.4 & 14.6 & 20.6 & 6.00 & 18.23 & 26.47 & 15.22 \\
& SU-Cos & 3.0 & 10.4 & 13.2 & 6.19 & 14.99 & 20.07 & 11.31 \\
& SU-MSE & 3.4 & 11.2 & 17.4 & 6.63 & 15.27 & 19.79 & 12.28 \\
\cmidrule[0.5pt]{2-9}
& \emph{AnyAttack-Cos}  & 3.8 & 14.0 & 22.8 & 7.36 & 20.71 & 27.55 & 16.04 \\
& \emph{AnyAttack-Bi}  & 4.8 &16.0 &23.6 &8.20 &22.31 &29.11 &17.34 \\
& \emph{AnyAttack-Cos w/ Aux} &  \underline{9.4} & \underline{24.6} & \underline{37.0} & \underline{11.51} & \underline{32.62} & \underline{48.18} & \underline{27.22} \\
& \emph{AnyAttack-Bi w/ Aux} &  \textbf{12.0} & \textbf{34.0} & \textbf{47.4} & \textbf{15.67} & \textbf{39.34} & \textbf{53.54} & \textbf{33.66} \\
\midrule[1pt]
\multirow{10}{*}{\rotatebox{0}{\textbf{ViT-L/14 $\times$ 336}}} 
&  AttackVLM-ii & 0.2 & 0.6 & 1.6 & 0.16 & 1.12 & 2.04 & 0.95 \\
&  AttackVLM-it & 0.2 & 0.6 & 1.8 & 0.32 & 0.96 & 1.76 & 0.94 \\
& SASD-WS-Cos & 2.8 & 10.8 & 16.4 & 6.52 & 18.31 & 26.19 & 13.50 \\
& SASD-WS-MSE & 4.4 & 13.6 & 19.2 & 6.72 & 18.23 & 25.71 & 14.64 \\
& SU-Cos & 2.4 & 8.0 & 11.2 & 4.88 & 13.79 & 18.39 & 9.78 \\
& SU-MSE & 3.6 & 8.2 & 13.2 & 6.40 & 14.51 & 19.19 & 10.85 \\
\cmidrule[0.5pt]{2-9}
& \emph{AnyAttack-Cos}  & 4.6 & 11.0 & 16.6 & 5.96 & 17.67 & 24.23 & 13.34 \\
& \emph{AnyAttack-Bi}  & 3.6 & 14.4 & 19.0 & 7.64 & 19.79 & 26.83 & 15.21 \\
& \emph{AnyAttack-Cos w/ Aux} &  \underline{9.0} & \underline{23.2} & \underline{37.2} & \underline{11.68} & \underline{34.03} & \underline{47.62} & \underline{27.12} \\
& \emph{AnyAttack-Bi w/ Aux} &  \textbf{12.0} & \textbf{33.2} & \textbf{46.8} & \textbf{14.79} & \textbf{39.22} & \textbf{53.06} & \textbf{33.18} \\
\bottomrule[1pt]
\end{tabular}
\end{adjustbox}
\caption{The retrieval performances on the MSCOCO dataset under different attacks. TR@1, TR@5, and TR@10 measures text retrieval performance, while IR@1, IR@5, and IR@10 measures image retrieval performance. R@Mean is the average of all retrieval metrics. Our proposed methods are \emph{italicized}, the best results are highlighted in \textbf{bold}, and the second-best results are \underline{underlined}.}
\label{tab:retrieval_mscoco}
\end{table*}

\begin{table}[t]
    \centering
    \small
    \begin{adjustbox}{max width=\textwidth}
    \begin{tabular}{@{}lr@{}}
        \toprule
        \textbf{Attack Method} & \textbf{Accuracy} \\
        \midrule
         AttackVLM-ii & 6.5 \\
         AttackVLM-it & 6.3 \\
        SASD-WS-Cos & 24.3 \\
        SASD-WS-MSE & 24.8 \\
        SU-Cos & 13.7 \\
        SU-MSE & 13.6 \\
        \cmidrule[0.5pt]{1-2}
        \emph{AnyAttack-Cos} & 17.5 \\
        \emph{AnyAttack-Cos w/ Aux}  & \textbf{44.8} \\
        \bottomrule
    \end{tabular}
    \end{adjustbox}
\caption{Attack performance comparison on the SNLI-VE dataset for multimodal classification.}
    \label{tab:classification}
\end{table}

\begin{table*}[t]
    \centering
    \small
    \setlength{\aboverulesep}{0pt} 
    \setlength{\belowrulesep}{0pt}
    \setlength{\extrarowheight}{0pt} 
    \begin{adjustbox}{max width=\textwidth}
    
    \begin{tabular}{@{}clcccccc@{}}
        \toprule
        Model & Attack Method & SPICE & BLEU-1 & BLEU-4 & METEOR & ROUGE-L & CIDEr \\
        \midrule
        \multirow{8}{*}{\rotatebox[origin=c]{0}{\textbf{InstructBLIP}}} 
        & AttackVLM-ii & 1.4 & 38.9 & 5.4 & 8.7 & 28.6 & 3.4 \\
        & AttackVLM-it & 1.3 & 39.1 & 5.4 & 8.7 & 28.8 & 3.3 \\
        & SASD-WS-Cos  & 3.4 & 43.9 & 7.2 & 10.5 & 32.2 & 10.9 \\
        & SASD-WS-MSE  & 3.2 & 44.6 & 7.0 & 10.8 & 32.4 & 11.8 \\
        & SU-Cos & 1.9 & 40.7 & 6.0 & 9.3 & 29.9 & 5.3 \\
        & SU-MSE & 1.9 & 40.9 & 6.5 & 9.5 & 29.9 & 6.0 \\
        \cmidrule[0.5pt]{2-8}
        & \emph{AnyAttack-Cos}  & 2.3 & 41.5 & 5.9 & 9.5 & 30.2 & 7.0 \\
        & \emph{AnyAttack-Cos w/ Aux} & \textbf{4.7} & \textbf{46.5} & \textbf{7.5} & \textbf{12.2} & \textbf{33.6} & \textbf{20.3} \\
        \midrule
        \multirow{8}{*}{\rotatebox[origin=c]{0}{\textbf{BLIP2}}} 
        & AttackVLM-ii & 1.2 & 39.6 & 5.3 & 8.7 & 29.0 & 3.6 \\
        & AttackVLM-it & 1.2 & 39.6 & 5.4 & 8.7 & 29.3 & 3.5 \\
        & SASD-WS-Cos & 2.6 & 43.0 & 6.3 & 10.2 & 32.0 & 9.3 \\
        & SASD-WS-MSE & 2.8 & 42.8 & 6.5 & 10.2 & 31.7 & 9.5 \\
        & SU-Cos & 1.6 & 40.9 & 5.6 & 9.2 & 30.1 & 4.7 \\
        & SU-MSE & 1.6 & 40.8 & 5.9 & 9.2 & 30.1 & 5.0 \\
        \cmidrule[0.5pt]{2-8}
        & \emph{AnyAttack-Cos}  & 1.8 & 41.3 & 5.2 & 9.6 & 30.9 & 5.6 \\
        & \emph{AnyAttack-Cos w/ Aux} & \textbf{3.3} & \textbf{44.2} & \textbf{6.0} & \textbf{11.0} & \textbf{32.4} & \textbf{13.3} \\
        \midrule
        \multirow{8}{*}{\rotatebox[origin=c]{0}{\textbf{BLIP}}} 
        & AttackVLM-ii & 1.3 & 39.8 & 5.0 & 8.8 & 29.9 & 3.4 \\
        & AttackVLM-it & 1.2 & 39.7 & 4.8 & 8.7 & 29.7 & 3.2 \\
        & SASD-WS-Cos & 3.3 & 43.8 & 6.9 & 10.7 & 32.3 & 11.9 \\
        & SASD-WS-MSE & \textbf{3.4} & 43.8 & 6.9 & 10.8 & 32.3 & 12.4 \\
        & SU-Cos & 2.6 & 43.0 & 6.5 & 10.1 & 31.8 & 8.7 \\
        & SU-MSE & 2.6 & 42.4 & 6.4 & 9.9 & 31.6 & 8.4 \\
        \cmidrule[0.5pt]{2-8}
        & \emph{AnyAttack-Cos}  & 2.2 & 41.6 & 6.0 & 9.5 & 31.1 & 6.1 \\
        & \emph{AnyAttack-Cos w/ Aux} & \textbf{3.4} & \textbf{44.4} & \textbf{7.1} & \textbf{11.1} & \textbf{32.8} & \textbf{13.0} \\
        \midrule
        \multirow{8}{*}{\rotatebox[origin=c]{0}{\textbf{Mini-GPT4}}} 
        & AttackVLM-ii & 1.6 & 29.5 & 2.3 & 9.3 & 24.3 & 1.6 \\
        & AttackVLM-it & 1.5 & 29.2 & 2.3 & 9.4 & 24.5 & 1.5 \\
        & SASD-WS-Cos & 2.8 & 30.5 & 2.4 & 10.9 & 25.2 & 2.6 \\
        & SASD-WS-MSE & 3.1 & 30.5 & 2.9 & 10.9 & 25.7 & 2.8 \\
        & SU-Cos & 2.0 & 29.5 & 2.9 & 9.9 & 24.8 & 1.9 \\
        & SU-MSE & 2.2 & 30.3 & 2.9 & 9.9 & 25.1 & 2.2 \\
        \cmidrule[0.5pt]{2-8}
        & \emph{AnyAttack-Cos}  & 2.5 & 30.5 & 2.4 & 10.3 & 25.2 & 1.9 \\
        & \emph{AnyAttack-Cos w/ Aux} & \textbf{4.6} & \textbf{32.5} & \textbf{4.0} & \textbf{12.4} & \textbf{27.3} & \textbf{5.3} \\
        \bottomrule
    \end{tabular}
    \end{adjustbox}
\caption{Attack performance comparison on the MSCOCO dataset for image captioning task.}\label{tab:caption}
\vspace{-10pt}
\end{table*}

In this section, we evaluate the performance of our proposed attack across multiple datasets, tasks, and VLMs. We evaluate the effectiveness of targeted adversarial attacks first in image-text retrieval tasks, then multimodal classification tasks, and finally image captioning tasks. Additionally, we analyze the performance of targeted adversarial images on commercial VLMs.

\subsection{Experimental Setup}

\paragraph{Baselines.} 

We first employed state-of-the-art targeted adversarial attack on VLMs, AttackVLM~\citep{zhao2024evaluating}. This method includes two variations: AttackVLM-ii and AttackVLM-it, which are based on different attack objectives. Both methods utilize the CLIP ViT-B/32 image encoder as the surrogate model, consistent with our approach.
Additionally, we incorporated two targeted adversarial attacks designed for visual classification models: SU~\citep{wei2023enhancing} and SASD-WS~\citep{wu2024improving}. 
Since the original cross-entropy loss used in these methods is not suitable for vision-language tasks, we modified them to employ cosine loss and mean squared error (MSE) loss to match targeted images. These modified methods are denoted as SU-Cos/SASD-WS-Cos and SU-MSE/SASD-WS-MSE, respectively.
For the SU attack, the surrogate model is CLIP ViT-B/32. 
For the SASD-WS attack, we utilized the officially released weights, as its surrogate model with the auxiliary model.
We denote our proposed methods as AnyAttack-Cos, AnyAttack-Bi, AnyAttack-Cos w/ Aux, and AnyAttack-Bi w/ Aux. These represent AnyAttack fine-tuned with \( \mathcal{L}_\text{Cos} \), fine-tuned with \( \mathcal{L}_\text{Bi} \), fine-tuned with \( \mathcal{L}_\text{Cos} \) using auxiliary models, and fine-tuned with \( \mathcal{L}_\text{Bi} \) using auxiliary models, respectively.

\paragraph{Datasets, Models, and Tasks.}

For the downstream datasets, we utilize the MSCOCO, Flickr30K, and SNLI-VE datasets. We employ a variety of target models, including CLIP, BLIP, BLIP2, InstructBLIP, and MiniGPT-4. The downstream tasks we focus on are image-text retrieval, multimodal classification, and image captioning. For each task, we selected the top 1,000 images. Additionally, following the methodology outlined in \citep{zhao2024evaluating}, we used the top 1,000 images from the ImageNet-1K validation set as clean (random) images to generate adversarial examples.

\begin{figure*}[htb] 
\centering
\includegraphics[width=\textwidth]{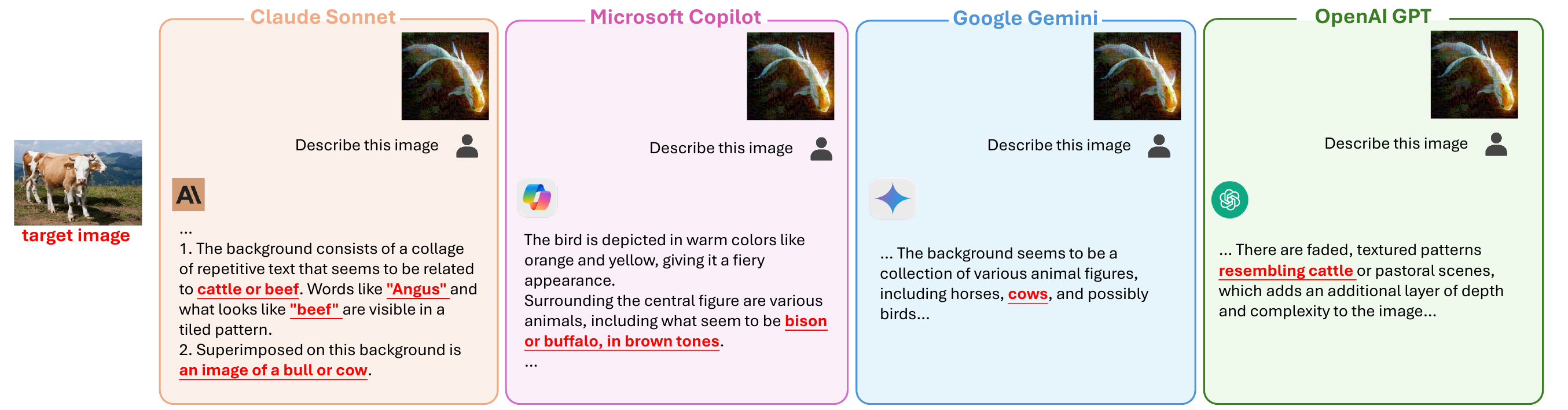} 
\caption{Example responses from commercial VLMs to targeted attacks generated by our method.} 
\label{fig:commercial} 
\vspace{-10pt}
\end{figure*}

\paragraph{Metric.}
We use the attack success rate (ASR) as the primary evaluation metric to assess the performance of targeted adversarial attacks. The calculation of ASR varies slightly depending on the specific task. For instance, in image-text retrieval tasks, ASR represents the recall rate between adversarial images and their corresponding ground-truth text descriptions. In multimodal classification tasks, ASR refers to the accuracy of correctly classifying pairs of ``adversarial image and ground-truth description."

\paragraph{Implementation Details.}
In this work, we study perturbations constrained by the \( \ell_\infty \) norm, ensuring that \( \| \delta \|_\infty \leq \epsilon \), where \( \epsilon \) represents the maximum allowable perturbation, set to \( \frac{16}{255} \).
We pre-trained the decoder for 520,000 steps on the LAION-400M dataset~\citep{schuhmann2021laion}, using a batch size of 600 per GPU on three NVIDIA A100 80GB GPUs. The optimizer used was AdamW, with an initial learning rate of \( 1 \times 10^{-4} \), which was adjusted using cosine annealing. For the downstream datasets, we fine-tuned the decoder for 20 epochs using the same optimizer, initial learning rate, and cosine annealing schedule.
We deployed two auxiliary models: the ViT-B/16 trained from scratch on ImageNet-1K and the ViT-L/14 EVA model~\citep{fang2023eva, fang2024eva}, both of which are trained on ImageNet-1K.
The factor \( K \) was set to 5. In the pre-training stage, the initial temperature \( \tau_0 \) was set to 1, the final temperature \( \tau_{\text{final}} \) was set to 0.07, and the total steps \( T \) were set to 10,000. 
More details can be found in the Appendix.

\subsection{Evaluation on Image-Text Retrieval}

In this subsection, we compare the performance of our method against baseline approaches on the image-text retrieval task. \cref{tab:retrieval_mscoco} presents the results on the MSCOCO dataset, while results on the Flickr30K dataset are detailed in Appendix. The following key observations can be made:

\begin{itemize}[left=0pt]
\item \textbf{Performance of AnyAttack-Bi w/ Auxiliary}: This variant achieves significantly superior performance compared to all baselines, surpassing the best-performing baseline by 15.02\%, 18.44\%, and 18.54\% on ViT-B/16, ViT-B/32, and ViT-L/14, respectively. All AnyAttack methods consistently deliver competitive results, outperforming most baselines. This highlights the effectiveness of our proposed method.

 \item \textbf{Effectiveness of the Auxiliary Module}: The Auxiliary module demonstrates its effectiveness, providing improvements of 6.455\%, 13.75\%, and 15.875\% on ViT-B/16, ViT-B/32, and ViT-L/14, respectively, when comparing AnyAttack w/ Auxiliary to AnyAttack.

\item \textbf{Advantages of Bidirectional Loss}: The bidirectional contrastive loss \( \mathcal{L}_{\text{Bi}} \) shows clear advantages for retrieval tasks, with AnyAttack-Bi consistently outperforming AnyAttack-Cos.
\end{itemize}

\subsection{Evaluation on Multimodal Classification}

Here, we compare the performance of our attack with the baselines on the multimodal classification task. \cref{tab:classification} presents the results on the SNLI-VE dataset. Our method, AnyAttack-Cos w/ Auxiliary, achieves the highest performance, surpassing the strongest baseline, SASD-WS-MSE, by 20.0\%. This underscores the effectiveness of our attack in multimodal classification tasks.

\subsection{Evaluation on Image Captioning}

Here, we evaluate the performance of our attack on the image captioning task using the MSCOCO dataset. The VLMs take adversarial images as input and generate text descriptions, which are then assessed against the ground-truth captions using standard metrics. \cref{tab:caption} presents the results across four VLMs: InstructBLIP, BLIP2, BLIP, and MiniGPT-4. Our proposed attack method, AnyAttack-Cos w/ Auxiliary, consistently demonstrates superior performance across all evaluation metrics, outperforming the baseline attacks on each VLM.

\begin{table}[t]
    \centering
    \small
    \begin{adjustbox}{max width=\textwidth}
    \begin{tabular}{lcc}
        \toprule
        \textbf{Attack Method} & \textbf{Google Gemini} & \textbf{OpenAI GPT} \\
        \midrule
         AttackVLM-ii & 0 & 2 \\
         AttackVLM-it & 0 & 0 \\
        SASD-WS-Cos & 5 & 28 \\
        SASD-WS-MSE & 1 & 25 \\
        SU-Cos & 0 & 11\\
        SU-MSE & 0 & 12\\
        \cmidrule[0.5pt]{1-3}
        \emph{AnyAttack-Cos w/ Aux}  & \textbf{31} & \textbf{38} \\
        \bottomrule
    \end{tabular}
    \end{adjustbox}
    \caption{Quantitative performance comparison on commercial VLMs. The reported values represent the ASR.}
    \label{tab:commercial}
    \vspace{-10pt}
\end{table}

\begin{figure*}[htb]
\centering
\includegraphics[width=\textwidth]{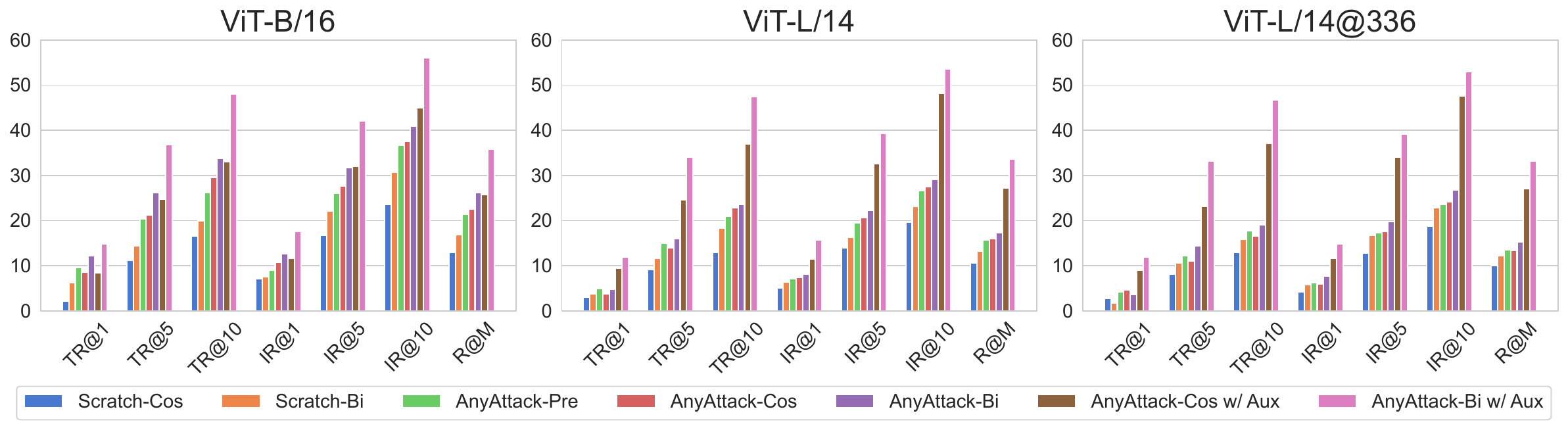} 
\caption{Performance comparison between different configurations of AnyAttack for the image-text retrieval task on MSCOCO. The plot shows the comparative performance of decoder initialized from scratch (Scratch), pre-trained (Pre), and fine-tuned (Cos and Bi), alongside the impact of auxiliary models (w/ Aux) and different fine-tuning objectives (Cos or Bi) on retrieval tasks.} 
\label{fig:ablation} 
\vspace{-10pt}
\end{figure*}

\subsection{Transfer to Commercial VLMs}

We evaluate the transferability of adversarial images generated by our method to commercial VLMs: Google Gemini, Claude Sonnet, OpenAI GPT, and Microsoft Copilot. Detailed setups are provided in Appendix.

\paragraph{Quantitative Results.}

We selected 100 images from the MSCOCO dataset as target images and conducted a comparison between our method and baseline approaches. The experiments were conducted using Google Gemini (Gemini 1.5 Flash) and OpenAI GPT (GPT-4o mini) via their respective APIs. Using the prompt ``Evaluate the relationship between the given image and text", along with corresponding options ``A) The text is highly relevant to the image. B) The text is partially relevant to the image. C) The text is not relevant to the image." \cref{tab:commercial} reports the percentage of responses labeled as ``highly and partially relevant" by these commercial VLMs. Our method consistently outperforms the baselines, achieving substantial improvements across all evaluated models.

\paragraph{Qualitative Results.}

To further demonstrate the effectiveness of AnyAttack, we uploaded the adversarial images to the publicly available web interfaces of these commercial VLMs. \cref{fig:commercial} showcases representative examples, with additional instances provided in Appendix. The portions of the VLM responses highlighted in red correspond to the target images, clearly illustrating the impact of our attacks on these VLMs.

\subsection{Further Analysis}

\begin{figure}[b]
    \centering
    \includegraphics[width=0.8\linewidth]{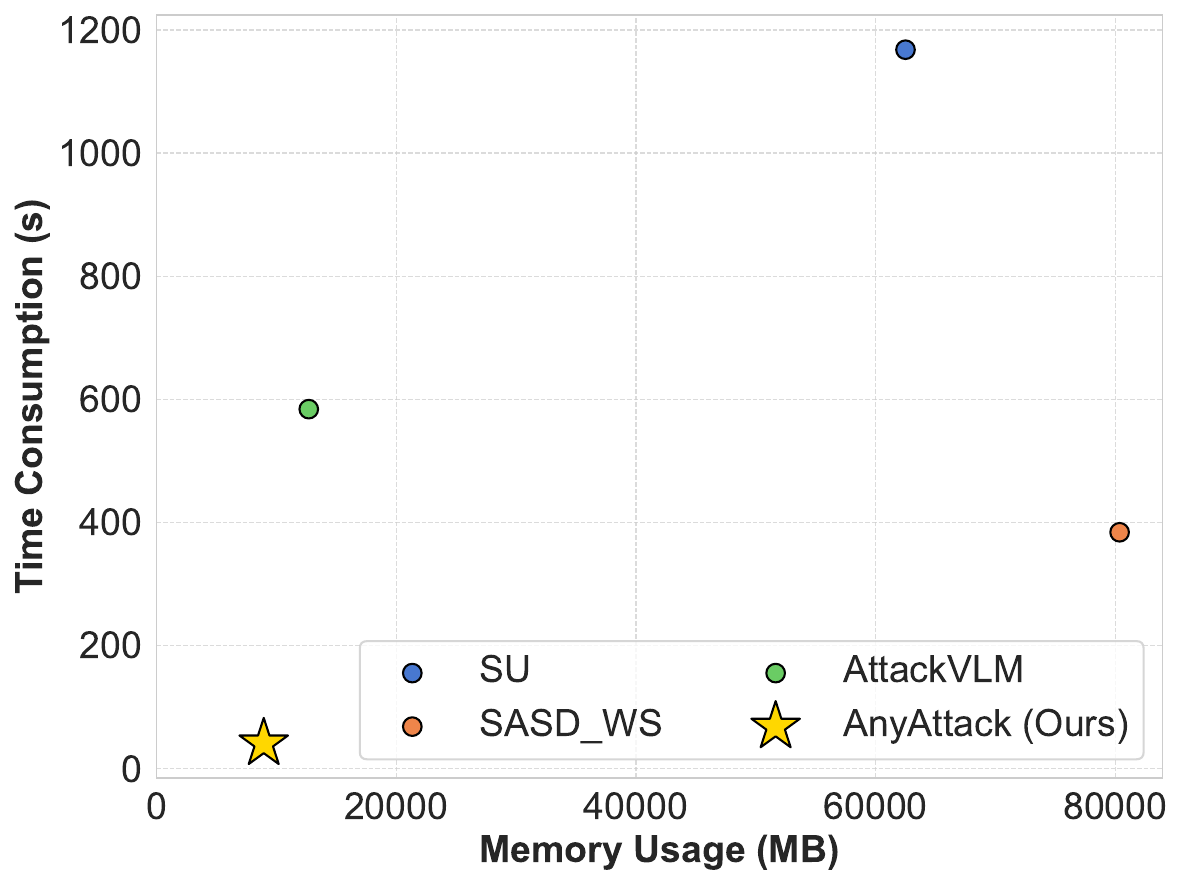}  
    \caption{Comparison of memory usage and time consumption across different methods.}
    \label{fig:efficiency_analysis}
    \vspace{-10pt}
\end{figure}

\paragraph{Ablation Study.}
We perform an ablation study on the MSCOCO dataset for image-text retrieval task to evaluate the impact of three key components in our approach: 1) \textbf{Training approach}: Pre-trained, fine-tuned, or trained from scratch. 2) \textbf{Auxiliary models}: With or without auxiliary model integration. 3) \textbf{Fine-tuning objective}: Cosine similarity ($\mathcal{L}_{\text{Cos}}$) vs. bidirectional contrastive loss ($\mathcal{L}_{\text{Bi}}$).

The results, summarized in \cref{fig:ablation}, reveal the following:
1) \textbf{Training approach}: Fine-tuning a pre-trained model achieves the highest performance, while training from scratch yields significantly worse results, indicating that pre-training is critical for task adaptation.
2) \textbf{Auxiliary models}: The inclusion of auxiliary models consistently improves performance, highlighting their role in enhancing transferability.
3) \textbf{Fine-tuning objective}: The bidirectional contrastive loss ($\mathcal{L}_{\text{Bi}}$) consistently outperforms the cosine similarity loss ($\mathcal{L}_{\text{Cos}}$), demonstrating its effectiveness in improving the alignment of image and text embeddings.

\paragraph{Efficiency Analysis.}
In this subsection, we compare the efficiency of our method with SU, SASD, and AttackVLM. Figure 5 presents the results for generating 1,000 adversarial images on a single NVIDIA A100 80GB GPU with a batch size of 250, showing both memory usage and time consumption. The results demonstrate that our approach significantly outperforms the baselines in both computational speed and memory efficiency.

\section{Conclusion}

In this paper, we introduced \textbf{AnyAttack}, a novel self-supervised framework for generating targeted adversarial attacks on VLMs. Our approach overcomes the scalability limitations of previous methods by enabling the use of \textbf{any} image to serve as a target for \textbf{attack} target without label supervision. Through extensive experiments, we demonstrated the effectiveness of AnyAttack across multiple VLMs and vision-language tasks, revealing significant vulnerabilities in state-of-the-art models. Our method showed considerable transferability, even to commercial VLMs, highlighting the broad implications of our findings.

These results underscore the urgent need for robust defense mechanisms in VLM systems. As VLMs become increasingly prevalent in real-world applications, our work opens new avenues for research in VLM security, particularly considering that this is the first time pre-training has been conducted on a large-scale dataset like LAION-400M.
This emphasizes the critical importance of addressing these challenges. 
Future work should focus on developing resilient VLMs and exploring potential mitigation strategies against such targeted attacks.

\section{Acknowledgements}

This work is supported by the National Key R\&D Program of China (Grant No. 2023YFC3310700, 2022ZD0160103) and the National Natural Science Foundation of China (Grant No. 62276067). This research has been made possible by a Research Impact Fund project (RIF R6003-21) and a General Research Fund project (GRF 16203224) funded by the Research Grants Council (RGC) of the Hong Kong Government.

{
    \small
    \bibliographystyle{ieeenat_fullname}
    \bibliography{main}
}


\end{document}